\documentclass{article} % For LaTeX2e
\usepackage[accepted]{icml2019}

\usepackage{amsmath,amsfonts,bm}

\def\eqref#1{equation~\ref{#1}}
\def\1{\bm{1}}

\DeclareMathAlphabet{\mathsfit}{\encodingdefault}{\sfdefault}{m}{sl}
\SetMathAlphabet{\mathsfit}{bold}{\encodingdefault}{\sfdefault}{bx}{n}

\newcommand{\E}{\mathbb{E}}

\newcommand{\R}{\mathbb{R}}

\DeclareMathOperator*{\argmax}{arg\,max}
\DeclareMathOperator*{\argmin}{arg\,min}

\usepackage{bm}
\usepackage[utf8]{inputenc} % allow utf-8 input
\usepackage[T1]{fontenc}    % use 8-bit T1 fonts
\usepackage{booktabs}       % professional-quality tables
\usepackage{amsmath,amssymb,amsfonts,amsthm}
\usepackage{mathtools}
\usepackage{nicefrac}       % compact symbols for 1/2, etc.
\usepackage{microtype}      % microtypography
\usepackage{graphicx}
\usepackage{subcaption}
\usepackage{dsfont}

\newcommand{\Y}{\mathcal{Y}}

\newcommand{\bx}{{x}}
\newcommand{\bz}{{z}}
\newcommand{\btheta}{{\theta}}

\newcommand{\cD}{\mathcal{D}}
\newcommand{\cH}{\mathcal{H}}
\newcommand{\cS}{\mathcal{S}}
\newcommand{\cT}{\mathcal{T}}
\newcommand{\cV}{\mathcal{V}}
\newcommand{\cW}{\mathcal{W}}

\newtheorem{theorem}{Theorem}

\newcount\comments
\newcommand{\hn}[1]{\ifnum\comments=1 {\color{green}{[HN: #1]}} \fi}
\newcommand{\highlight}[1]{\ifnum\comments=1 {\color{blue}{#1}} \else {#1} \fi}

\begin{document}

\twocolumn[
\icmltitle{Metric-Optimized Example Weights}
% Authors must not appear in the submitted version. They should be hidden
% as long as the \iclrfinalcopy macro remains commented out below.
% Non-anonymous submissions will be rejected without review.
\icmlsetsymbol{equal}{*}

\begin{icmlauthorlist}
\icmlauthor{Sen Zhao}{equal,google}
\icmlauthor{Mahdi Milani Fard}{equal,google}
\icmlauthor{Harikrishna Narasimhan}{google}
\icmlauthor{Maya Gupta}{google}
\end{icmlauthorlist}

\icmlaffiliation{google}{Google AI, 1600 Amphitheatre Parkway, Mountain View, CA 94043, USA}

\icmlcorrespondingauthor{Sen Zhao}{senzhao@google.com}

\vskip 0.3in
]

% The \author macro works with any number of authors. There are two commands
% used to separate the names and addresses of multiple authors: \And and \AND.
%
% Using \And between authors leaves it to \LaTeX{} to determine where to break
% the lines. Using \AND forces a linebreak at that point. So, if \LaTeX{}
% puts 3 of 4 authors names on the first line, and the last on the second
% line, try using \AND instead of \And before the third author name.

\newcommand{\fix}{\marginpar{FIX}}
\newcommand{\new}{\marginpar{NEW}}

%\iclrfinalcopy % Uncomment for camera-ready version, but NOT for submission.

\printAffiliationsAndNotice{\icmlEqualContribution} % otherwise use the standard text.

\begin{abstract}
Real-world machine learning applications often have complex test metrics, and may have training and test data that are not identically distributed. Motivated by known connections between complex test metrics and cost-weighted learning, we propose addressing these issues by using a weighted loss function with a standard loss, where the weights on the training examples are learned to optimize the test metric on a validation set. These metric-optimized example weights can be learned for any test metric, including black box and customized ones for specific applications. We illustrate the performance of the proposed method on diverse public benchmark datasets and real-world applications. We also provide a generalization bound for the method.
\end{abstract}

\section{Introduction}
\label{sec:intro}
In machine learning, each example is usually weighted equally during training. Such uniform weighting delivers satisfactory performance when training and test examples are independent and identically distributed (IID), and the training loss matches the test metric. However, these requirements are often violated in real-world applications, as the real-world goals for a model are often quite complicated. 
%For example, a test metric may be multi-objective,computed over quantiles or subsets of the data, and/or highly nonlinear like the win-loss ratio to an existing model.  
If the training loss does not correlate sufficiently with the test metric, inferior test performance can result \citep{Cortes:2003, Perlich:2003, Davis:2006}. 

If the test metric has a sufficiently nice structure, and the train and test distributions are IID, it may be possible to specify a useful example-weighted training loss, and train for it \cite{Koyejo+14, Parambath:2014, Narasimhan:2015a, Narasimhan:2015b}. In this paper, we extend the strategy of using training example weights to arbitrary test metrics, by learning a function to weight the examples that best optimizes the test metric on a validation set.  Our proposal, \emph{metric-optimized example weights (MOEW)} is applicable for \emph{any} test metric. MOEW can address non-IID test samples if there is available a small set of labeled validation examples that are IID with the test examples.  By learning a weighting function, MOEW effectively rescales the loss on each training example, and reshapes the overall loss such that its optima better match the optima of the test metric.

\begin{figure}[ht]
\vspace{-7pt}
\centering
    \begin{subfigure}{.47\linewidth}
        \includegraphics[width=\linewidth]{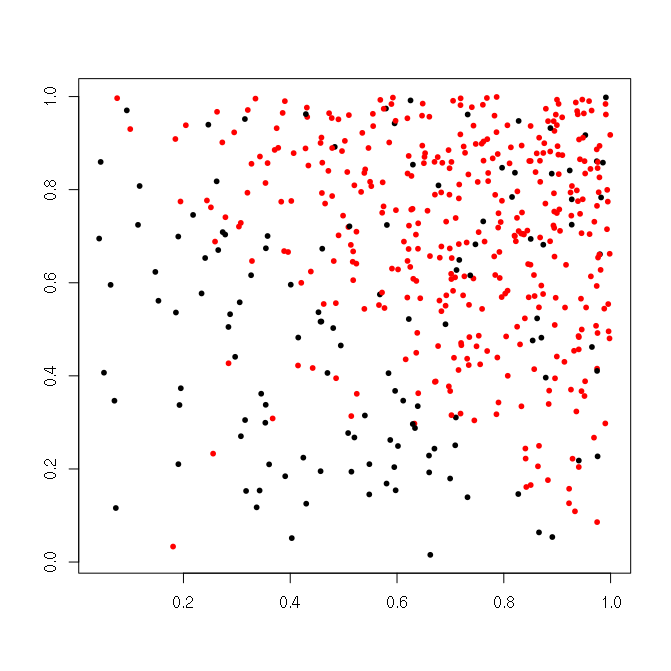}
        \vspace{-2.5em}
        \caption{Training Distribution}\label{fig:fig_a}
    \end{subfigure} %
    \begin{subfigure}{.47\linewidth}
        \includegraphics[width=\linewidth]{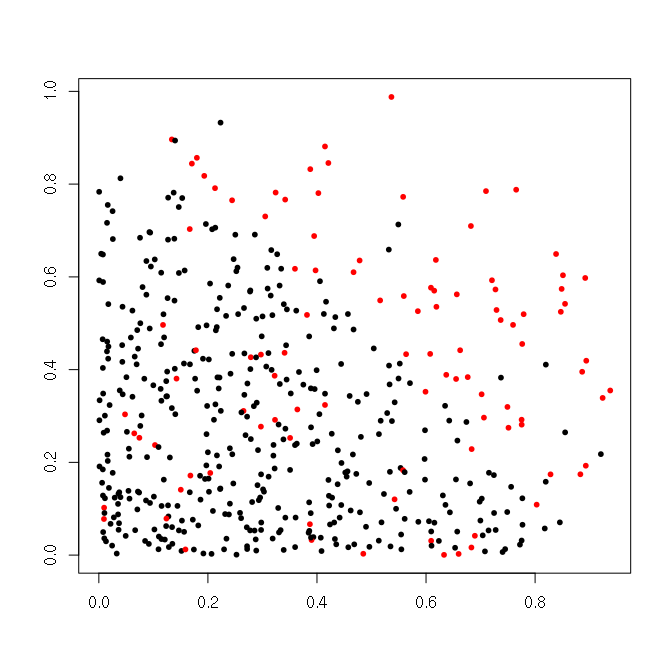}
        \vspace{-2.5em}
        \caption{Test Distribution}\label{fig:fig_b}
    \end{subfigure} %
    \begin{subfigure}{.47\linewidth}
        \includegraphics[width=\linewidth]{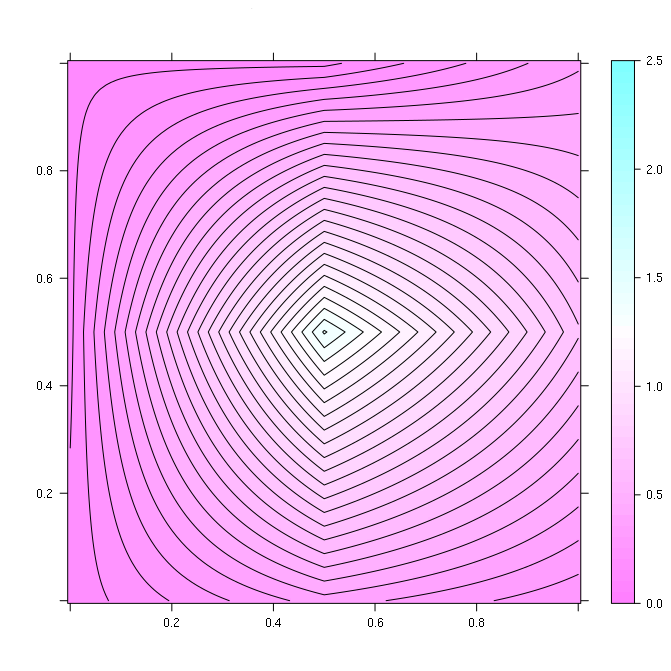}
        \vspace{-2.5em}
        \caption{Pos. Example Weighting}\label{fig:fig_c}
    \end{subfigure}
    \begin{subfigure}{.47\linewidth}
        \includegraphics[width=\linewidth]{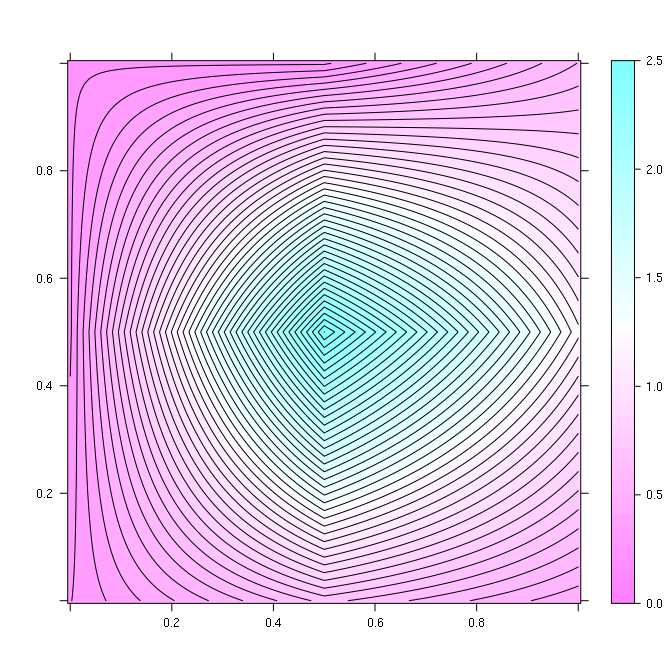}
        \vspace{-2.5em}
        \caption{Neg. Example Weighting}\label{fig:fig_d}
    \end{subfigure}
\vspace{-3pt}
\caption{Figures \ref{fig:fig_a} and \ref{fig:fig_b} show the distribution of the training and validation/test data: black dots are negative examples and red dots are positive examples. The true decision boundary is the diagonal of the square from the upper-left to lower-right corner. The two features were drawn from beta distributions:  training $(x_1, x_2)\sim(\beta(2, 1), \beta(2, 1))$, and validation/test $(x_1, x_2)\sim(\beta(1, 2), \beta(1, 2))$.   Figures \ref{fig:fig_c} and \ref{fig:fig_d} show contour lines of the MOEW weighting function for positive and negative examples, respectively.
\vspace{-0.5cm}
}\label{fig:toy}
\end{figure}

As an illustrative example, Figure~\ref{fig:toy} shows a simulated dataset with non-IID training and test examples, and the learned MOEW example weighting.  The goal is to maximize precision at 95\% recall on the test distribution. At 95\% recall, with uniform weighting the precision is 20.8\%; with optimal importance weighting \citep{Shimodaira:2000} it is better at 21.8\%, but with MOEW it can be improved to 23.2\%, much closer to the Bayes optimal of 25\%. Comparing Figure \ref{fig:fig_c} to \ref{fig:fig_d}, one can see that MOEW learns to upweight negative training examples compared to positive examples, and to upweight examples closer to the center. 

\section{Related Work}
\label{sec:relatedworks}
Our proposed approach simultaneously addresses two key issues: 
% non-constant noise levels across examples, 
non-identically distributed training and test sets, and
mismatch between training and test metrics.  
%Several existing methods summarized below address some of these issues, but none of the methods %provide a method to 
% address all of them at once.

% Maximum likelihood based inference is commonly used to address the issue with non-constant noise levels in the training data. In addition, several common machine learning techniques, such as ensemble and graph-based methods, are found to be robust to binary label noise; see \citet{Frenay:2014} for a survey of these methods.

% For the first issue, maximum likelihood based inference is commonly used to address the issue of non-constant noise levels in the training data.

For the first issue, classical approaches include propensity score matching \citep{Lunceford:2004} and importance weighting \citep{Shimodaira:2000, Sugiyama:2007, Sugiyama:2008, Kanamori:2009}. %These techniques can also be adapted for model selection \citep{Sugiyama:2007}
There has also been work on discriminative approaches for learning under  covariate shift \cite{Bickel:2009}. Other related work include methods that are robust to label noise in training data (see e.g.\ \citet{Frenay:2014}) and maximum likelihood inference to deal with non-constant noise levels in data. None of these approaches handle a general test metric.

For the second issue, a variety of approaches have been developed to directly optimize complex metrics on the training set. These include 
(i) \textit{plug-in} approaches \cite{Koyejo+14, Narasimhan:2014}, (ii) \textit{surrogate loss optimization} for
AUC \citep{Ferri:2002, Yan:2003, Cortes:2003, Freund:2003, Herschtal:2004, Rudin:2009, Zhao:2011}, the F-measure \citep{Joachims:2005, Jansche:2005} and other ranking metrics \citep{Yue:2007, Eban:2017, Kar:2014, Kar:2015}, 
and (iii) \textit{sequential example-weighting} techniques for complex metrics \cite{Parambath:2014, Narasimhan:2015a, Narasimhan:2015b} and constraints \cite{Agarwal+18, Kearns+18, Narasimhan18}. These methods crucially rely on the test metric having a specific closed-form structure and do not offer the generality of our approach.
Of these, our approach is most closely related to the sequential weighting techniques, and generalizes the idea of example-weighting to arbitrary (possibly black-box) test metrics and to non-identical train and test distributions.

\section{Metric Adaptive Weight Optimization}\label{sec:proposal}
We now describe our proposed adaptive example weighting approach
% We propose learning an optimal example weighting function, trained on validation scores for different example weightings.  Our proposal is suitable for \emph{any test metrics}. 
and provide theoretical justifications in Section~\ref{sec:theory}.

\subsection{Overview}\label{sec:propoverview}

We consider classification and regression problems where examples $x \in \R^D$ and labels $y \in \Y \subseteq \R$. We allow the training and test distribution to be non-identical, but sharing the same support. This may be the case, for example, when the marginal distribution of the features $p(x)$ changes over time (e.g., in a cancer prediction task, there could be demographic changes in the patient population), while the conditional-label distribution $p(y|x)$ remains the same. 

We are interested in learning a classifier or regressor, $h(x;\btheta)\in\cH$ for $\btheta \in \R^D$, which is parameterized by $\btheta$. The performance of the classifier or regressor is evaluated using a metric $M(\theta)$ defined on a test distribution (with higher values of $M$ being better). 
Let $\cT$ and $\cV$ denote the sets of training and validation examples, respectively. We assume that $\cV$ is drawn IID from the test distribution, but that $\cT$ and $\cV$ may not be identically distributed. The goal is to use the training and validation sets to find model $\btheta^*$ that maximizes the test metric $M$.
% The goal is to find the model with the maximum test metric:
% \begin{eqnarray}
% \theta^* \in \argmax_{\theta \in \R^D}\, M(\theta).
% \label{eq:learning-problem}
% \end{eqnarray}

Let  $\hat M(x_\cV, y_\cV; \btheta)$ denote an estimate of the test metric on the validation dataset $\cV$. One could consider directly optimizing $\hat M(x_\cV, y_\cV; \btheta)$ over $\theta \in \R^D$. However, for a high-dimensional $\btheta$ and a small $\cV$, we may end up overfitting to the validation set, and performing poorly on the test metric.

On the other hand, it is known that for a wide range of evaluation metrics that can be written as a function of simpler expected point-wise losses, the optimal parameters $\theta$ can be found by minimizing a particular example-weighted loss function \cite{Parambath:2014, Narasimhan:2015b, Narasimhan:2015a} (we provide an example of one such family in Section \ref{sec:theory}).
Motivated by these results, we propose to learn a weighting function for the training examples that yields an optimal model for the test metric. 
%We minimize a weighted loss on the training set and use the  validation set to estimate the test metric for the obtained model parameters. 
%Unlike previous approaches, we only need blackbox access to $M$. %the test metric.
%and allow the training examples to be distributed differently from the test set. 

In particular, we use an example weighting function $w: \R^D \times \R \rightarrow \R_+$ parameterized by $\alpha$ that maps a training feature vector and label to a non-negative weight. Given a loss function $L(\hat y, y)$ between a predicted label $\hat{y}$ and true label $y$, we seek to find a model  $\hat{\theta}$ that minimizes this loss on the training examples, with each example weighted by $w$ (we use standard methods such as SGD for this optimization):
\begin{align}\label{eq:optimization}
    \hat \theta(\alpha) = \argmin_{\btheta} \sum_{j\in\cT}w\left(\bx_j,y_j;\alpha\right)  L\left(h\left(\bx_j;\btheta \right), y_j\right).
\end{align}
\vspace{-15pt}
Our proposed approach is to learn the example weighting function $w(x, y;\hat \alpha)$ that optimizes $\hat M$:
%the test metric on the validation set:
\begin{align}\label{eq:weight}
    \hat \alpha=\argmax_{\alpha} \hat M(x_\cV, y_\cV; \hat\btheta(\alpha)).
    \vspace{-8pt}
\end{align}
In other words, we propose finding the optimal parameters $\hat \alpha$ for the example weighting function $w(x, y;\hat \alpha)$ such that the model that optimizes the example-weighted loss on the training set achieves the best validation score. The benefit of this approach is that $\alpha$ can be constructed to be low-dimensional, which can be optimized to maximize $M$ on the validation set, whereas $\btheta$, being high-dimensional, is optimized on the larger training set. 

To simplify the notation, where possible, we shall denote $\btheta(\alpha)$ by $\btheta$ and $\hat M(x_\cV, y_\cV; \btheta)$ by $\hat M(\btheta)$. 
The validation metric $\hat M(\btheta)$ is likely non-convex and non-differentiable in $\theta$ and $\alpha$, which makes it hard to directly optimize $\hat M(\btheta)$ through, e.g.\ SGD. Instead, we adopt an iterative algorithm to optimize for $\btheta^*$ and $\hat \alpha$, which is detailed in Algorithm~\ref{alg:full}. 

We start with a random sample of $K$ weighting parameters, $\alpha^0=\{\alpha_1^0,\dots,\alpha_K^0\}$. For each of the weighting parameters $\alpha_l^0$, $1\leq l\leq K$, we solve \eqref{eq:optimization} to obtain the $K$ corresponding model parameters $\hat\btheta_l^0$, and use those to compute the $K$ validation metrics, $\hat M(\hat\btheta_l^0)$. Note this step can be parallelized. Then, based on the batch of $K$ weighting parameters and validation metrics, we determine a new set of $K$ weight parameter candidates. This step is performed by a call to the subroutine  \textit{GetCandidateAlphas}. The process is repeated for $B$ iterations. At the end, we choose the candidate $\alpha$ that produced the best validation metric.

In the following subsections, we describe the function class $\cW$ of the weighting model, and one possible instantiation of the \textit{GetCandidateAlphas} subroutine used in Algorithm~\ref{alg:full}.

\begin{algorithm}[tb]
\caption{Get optimal $\hat \alpha$ and $\hat \theta(\hat \alpha) $\label{alg:full}}  
    \begin{algorithmic} 
            \STATE $\cT \gets \text{training data}$
            \STATE $\cV \gets \text{validation data}$
            \STATE $B\in\mathbb{N}_+ \gets \text{number of batches of weight parameters}$
            \STATE $\alpha^0=\{\alpha_1^0,\dots,\alpha_K^0\} \gets \text{initial weight parameters}$
            \FOR{$i = 0, 1, \dots, B-1$}
                \FORALL{$l\in\{1,\dots,K\}$}
                    \STATE $\hat\btheta_l^i \gets \argmin_{\btheta} \sum_{j\in\cT} w(\bx_j,y_j;\alpha_l^i) L(h(\bx_j;\btheta), y_j)$
                   % \State $m_l^i \gets \hat M(\btheta_l^i)$ I don't think we need this extra notation
                \ENDFOR
                \STATE $s^i \gets \{(\alpha_1^i, \hat M(\hat\btheta_1^i)),\dots,(\alpha_K^i,\hat M(\hat\btheta_K^i))\}$
                \STATE $\alpha^{i+1} \gets \textit{GetCandidateAlphas}(s^0,\dots,s^i)$
            \ENDFOR
            \STATE $\hat b, \hat k \gets \argmax_{b\in\{0,\dots,B-1\}, k\in\{1,\dots,K\}} \hat M(\btheta_k^b)$
            \STATE $\hat \alpha, \hat \theta \gets \alpha_{\hat k}^{\hat b},\hat\theta_{\hat k}^{\hat b}$
    \end{algorithmic}  
\end{algorithm}

\subsection{Function Class for the Example Weighting Model}
Recall that the final optimal $\hat \alpha$ is taken to be the best out of $B\times K$ samples of $\alpha$'s. To ensure a sufficient coverage of the weight parameter space for a small number of $B\times K$ validation samples, we found it best to use a function class $\cW$ with a small number of parameters. 

There are many reasonable strategies for defining $\cW$. In this paper, we chose the functional form:
\begin{align} \label{eq:weightembedding}
    w\left(\bx,y;\alpha\right) = c \: \pi\left(y\right) \sigma\left(\bz(x,y)^\top\alpha\right),
\end{align}
where $c \in \R_+$ is a constant that normalizes the weights over (a batch from) the training set $\mathcal{T}$, and $\sigma(z(x,y)^\top\alpha)$ is a sigmoid transformation of a linear function of a low-dimensional embedding $z$ of $(x,y)$. In the experiments discussed in Section~\ref{sec:exps}, we used the standard importance function 
%\begin{align} \label{eq:importance}
%\pi\left(y\right)=\frac{p_{\cV}\left(y\right)}{p_{\cT}\left(y\right)},
%\end{align}
$\pi\left(y\right)=p_{\cV}\left(y\right)/p_{\cT}\left(y\right)$,
where $p_{\cV}(y)$ and $p_{\cT}(y)$ denote the probability density function of $y$ in the validation and training data, respectively. In practice, $\pi(y)$ can be substituted with any baseline weighting function, which can be considered as an initialization of MOEW.

While there are many ways to form a low-dimensional embedding $z$ of $(x,y)$, we choose to use an autoencoder. To train the autoencoder $z(x,y; \hat \xi)$, we minimize the weighted sum of the reconstruction loss for $x$ and $y$: 
\begin{align}\label{eq:autoencoder}
    \sum_{(x,y)\in\cT}
    \!\! \big\{
    \lambda\, L_x( z(x,y;\xi), x ) 
    + (1 - \lambda)\, L_y (z(x,y;\xi), y )
    \big\},\nonumber
\end{align}
where $L_x$ is an appropriate loss for the feature vector $x$ and $L_y$ is an appropriate loss for the label $y$. The hyperparameter $\lambda$ is used to adjust the relative importance of features and the label in the embedding. For all the experiments in this paper discussed in Section~\ref{sec:exps}, we used a fixed $\lambda = 0.5$.

\begin{algorithm}[tb]
\caption{Get Candidate $\alpha^{i+1}$\label{alg:candidate}}  
    \begin{algorithmic}
            \STATE $\cS \gets \text{set of $\alpha$ and validation metrics in prior batches}$
            \STATE $K\in\mathbb{N}_+ \gets \text{number of candidates in a batch}$
            \STATE $\mathds{B} \gets \text{convex domain in which to sample $\alpha$}$
            \STATE $p, q\in[0,100] \gets \text{explore-exploit hyperparameters}$
            \FOR{$j = 1, 2, \dots, K$}
                \STATE $g(\alpha) \gets \text{GPR}(\cS)$
                \STATE $\alpha_j^{i+1} \gets \argmax_{\alpha\in \mathds{B}}\{Q_{(50 + p/2)\%}[g(\alpha)]\}$
                \STATE $\cS \gets \cS \, \cup \, \{(\alpha_j^{i+1}, Q_{(50-q/2)\%}[g(\alpha_j^{i+1})])\}$
            \ENDFOR
            \STATE $\alpha^{i+1} \gets \{\alpha_1^{i+1},\dots,\alpha_K^{i+1}\}$
    \end{algorithmic}  
\end{algorithm}

\begin{figure}[ht]
\centering
    \vspace{-0.5em}
    \begin{subfigure}{.47\linewidth}
        \includegraphics[width=\linewidth]{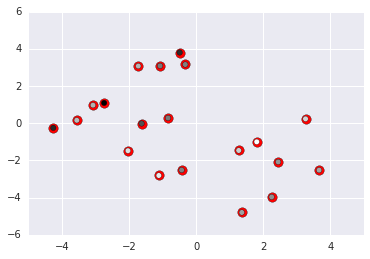}
        \vspace{-1.6em}
        \caption{Batch 0}\label{fig:GPR0}
    \end{subfigure} %
    \begin{subfigure}{.47\linewidth}
        \includegraphics[width=\linewidth]{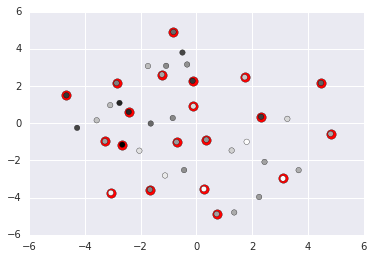}
        \vspace{-1.6em}
        \caption{Batch 1}\label{fig:GPR1}
    \end{subfigure} %
    \begin{subfigure}{.47\linewidth}
        \includegraphics[width=\linewidth]{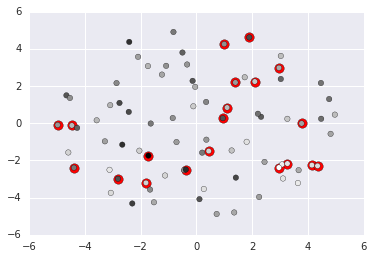}
        \vspace{-1.6em}
        \caption{Batch 3}\label{fig:GPR3}
    \end{subfigure} %
    \begin{subfigure}{.47\linewidth}
        \includegraphics[width=\linewidth]{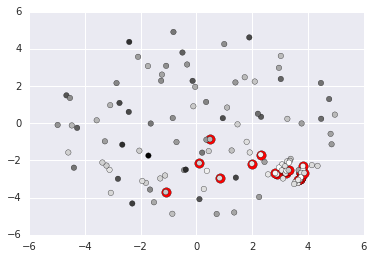}
        \vspace{-1.6em}
        \caption{Batch 5}\label{fig:GPR5}
    \end{subfigure} %
    \begin{subfigure}{.47\linewidth}
        \includegraphics[width=\linewidth]{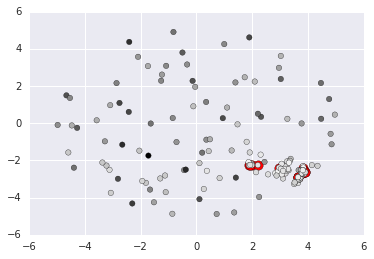}
        \vspace{-1.6em}
        \caption{Batch 7}\label{fig:GPR7}
    \end{subfigure}
    \begin{subfigure}{.47\linewidth}
        \includegraphics[width=\linewidth]{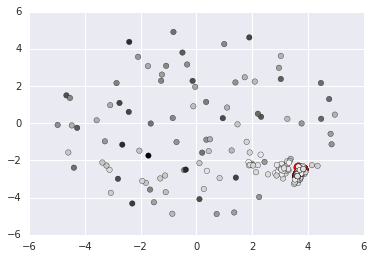}
        \vspace{-1.6em}
        \caption{Batch 9}\label{fig:GPR9}
    \end{subfigure}
    \begin{subfigure}{.47\linewidth}
        \includegraphics[width=\linewidth]{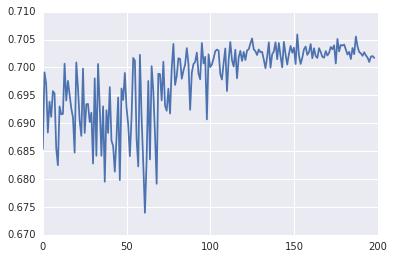}
        \vspace{-1.6em}
        \caption{Validation Metric}\label{fig:GPRvalidation}
    \end{subfigure}    
    \begin{subfigure}{.47\linewidth}
        \includegraphics[width=\linewidth]{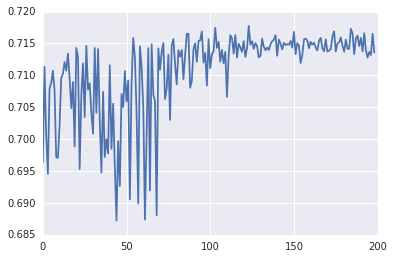}
        \vspace{-1.6em}
        \caption{Test Metric}\label{fig:GPRtest}
    \end{subfigure}
\caption{In Figures~\ref{fig:GPR0}-\ref{fig:GPR9}, red dots show the first two embedding dimensions of candidate $\alpha$'s throughout the $B=10$ batches GPR sampling process in the example studied in Section~\ref{sec:DSA}. The dots are shaded based on the magnitudes of $\alpha$'s. Figures~\ref{fig:GPRvalidation} and \ref{fig:GPRtest} show the validation and test metrics with the 200 sampled candidate $\alpha$. The correlation between validation (g) and test (h) metrics is 0.97.}\label{fig:GPR}
\vspace{-10pt}
\end{figure}

\subsection{Global Optimization of Weight Function}\label{sec:globaloptimization}
The validation metric $\hat M(\hat \theta(\alpha))$ may have multiple optima as a function of $\alpha$. In order to find the maximum validation score, the sampled candidate $\alpha$'s should achieve two goals. First, $\alpha$ should sufficiently cover the weighting parameter space. In addition, we also need a large number of candidate $\alpha$'s sampled near the most promising local optima. In other words, there is an exploration (spread $\alpha$'s more evenly) and exploitation (make $\alpha$'s closer to the optima) trade-off when choosing candidate $\alpha$'s.

One can treat this as a global optimization problem and sample candidate $\alpha$'s with a derivative free optimization algorithm \citep{Conn:2009}, such as simulated annealing \citep{vanLaarhoven:1987}, particle swarm optimization \citep{Kennedy:1995} and differential evolution \citep{Storn:1997}. For a low dimensional space, one can do an exhaustive search for $\alpha$'s on a grids. We derive a generalization bound for such a search in Theorem~\ref{thm:generic_bound}. 

For the experiments in this paper, we chose to base our algorithm on Gaussian process regression (GPR), specifically on the Gaussian Process Upper-Confidence-Bound (GP-UCB)~\citep{Auer:2002,AuerBianchi:2002} adapted to batched sampling of parameters, i.e., GP-BUCB. \citet{Desautels:2014} shows that GP-BUCB could achieve the same cumulative regret as GP-UCB up to a constant factor.

As detailed in Algorithm~\ref{alg:candidate}, after getting the $i$-th batch of candidate $\alpha$'s and their corresponding validation metrics $\hat M(\hat\btheta)$, we build a GPR model $g(\alpha)$ to fit the validation metrics on $\alpha$ for all previous observations. The next batch of candidate $\alpha$'s is then selected sequentially: we first sample an $\alpha_1^{i+1}$ based on the upper bound of the $p$\% prediction interval of $g(\alpha)$, i.e., $\alpha_1^{i+1}=\argmax_{\alpha} Q_{(50+p/2)\%}[g(\alpha)]$. A larger value of hyperparameter $p$ encourages exploration, whereas a smaller value encourages exploitation. After $\alpha_1^{i+1}$ is sampled, we refit a GPR model with an added observation for $\alpha_1^{i+1}$, as if we have observed a validation metric $Q_{(50-q/2)\%}[g(\alpha_1^{i+1})]$, which is the lower bound of the $q\%$ prediction interval of $g(\alpha)$. Hyperparameter $q$ controls how much the refitted GPR model trusts the old GPR model and a larger $q$ encourages wider exploration within each batch. We then use the refitted GPR model to generate another candidate $\alpha_2^{i+1}$, and continue this process until all the candidate $\alpha$'s in the $(i+1)$-th batch are generated. Note that to ensure convergence, in practice, we usually generate candidate $\alpha$'s within a bounded domain $\mathds{B}$.

%With GPR models, we can stop the candidate generation process early if the width of the prediction interval throughout the bounded space is smaller than a certain threshold.

%In the experiments presented in Section~\ref{sec:exps}, $p$ and $q$ were set so that the $(50+p/2)\%$ and $(50-q/2)\%$ standard Gaussian quantile values equal to $\pm1$, i.e., $p=q=68.3$. 
Figure~\ref{fig:GPR} shows 200 sampled candidate $\alpha$'s in $B=10$ batches in the example in Section~\ref{sec:DSA} with $p=q=68.3$. It shows that at the beginning of the candidate $\alpha$ sampling process, GPR explores more evenly across the domain $\mathds{B}$. As the sampling process continues, GPR begins to exploit more heavily near the optimal $\hat \alpha$ in the lower right corner.

%We can also stop the candidate generation process early if the width of the prediction interval throughout the bounded space $\cD$ is smaller than a certain threshold. 

%\subsection{Get the Optimal Alpha Star}
%We propose two approaches to get the optimal $\hat \alpha$ from a set of candidate $\alpha$'s and their corresponding validation metrics. The na\"ive approach is to simply take the candidate $\alpha$ with the maximum validation metric as $\hat \alpha$. In most of the cases we have tested, this na\"ive approach works well, especially when the validation metric is stable with a low variance. 

%However, if the validation data are small, or if the metric only depends on a few examples in the validation data (such as precision at high decision threshold), the validation metric is usually noisy. To address this issue, instead of using the na\"ive approach, we can build another GPR model $g^*(\alpha)$ to fit the validation metric $M[h(\bx_{\cV};\btheta),y_{\cV}]$ as a function of the candidate $\alpha$. We then obtain $\hat \alpha=\argmax_{\alpha\in\cD}\{g^*(\alpha)\}$ based on $g^*(\alpha)$. The predicted value in $g^*(\alpha)$ depends on all of the candidate $\alpha$'s, which reduce the variance in $\hat \alpha$.

\section{Theoretical Analysis}
\label{sec:theory}

As a motivation for the MOEW approach, 
we first present a family of evaluation metrics for which the optimal model can be found through  example-weighting.
We then prove a generalization bound for an instance of the MOEW algorithm. All proofs are provided in the appendix.

Let $\cD$ and $\cD'$ denote the test and train distributions respectively. We consider evaluation metrics $M(\theta)$ of the form:
\begin{eqnarray}
M(\theta) &\coloneqq& \Psi(\phi_1(\theta), \ldots, \phi_K(\theta)), \label{eqn:metric} %\\
\end{eqnarray}
where
$\Psi: \R^K \rightarrow \R$ is a general function of $K$ simpler  metrics $\phi_k$ defined on the \emph{test} distribution:
\begin{eqnarray*}
\phi_k(\theta) &\coloneqq& \E_{(x,y) \sim \cD}\big[\ell_k\big(x, y; \theta \big)\big],
\end{eqnarray*}
with each $\ell_k: \R^D \times \Y \rightarrow \R_+$ being a loss on individual examples. This class of metrics includes the F-measure, various fairness metrics and general functions of the confusion matrix of a classifier. %In fact, similar functional forms have been used previously to model arbitrary functions on a set of points \cite{Zaheer+17}.
As a simple example, to write the \emph{precision} of a classifier in this form, we can set:
\begin{eqnarray*}
\ell_1(x, y; \theta) &\coloneqq& \1(h(x; \theta) \!=\! 1, y \!=\! 0) \nonumber \\
\ell_2(x, y; \theta) &\coloneqq& \1(h(x; \theta) \!=\! 0, y \!=\! 1) \nonumber \\
\Psi(\phi_1, \phi_2) &\coloneqq& (\E[y \!=\! 1] - \phi_2) / (\E[y \!=\! 1] + \phi_1 - \phi_2).  \nonumber
\end{eqnarray*}

\subsection{Optimality of Example-weighting}
We show that under a specific set of assumptions on $\ell_k$, the optimal model for the above test metrics can be found by minimizing an example-weighted loss on the \textit{train distribution}. This result is based on and generalizes similar results from \citep{Narasimhan:2015b, Narasimhan:2015a}.

\begin{theorem}
Let $\theta^*$ be the optimal model parameters for $M(\theta)$. Suppose the train and test distributions $\cD$ and $\cD'$ have the same support. Suppose each $\phi_k$ is strictly convex in $\theta$  and each $\ell_k\big(x, y; \theta \big) \,=\, \big(\varphi_{k}(x, y)\big)^\top L\big(h(x; \theta), y\big)$, where $L: \R \times \Y \rightarrow \R_+^m$ is a $m$-dimensional loss function, and $\varphi_{1}, \ldots, \varphi_{K}: \R^D \times \Y \rightarrow \R_+^m$ are $m$-dimensional cost functions. Suppose $\Psi$ satisfies \textbf{one} of the following:
\begin{enumerate}
    \item $\Psi(z)$ is concave in $z$, and is strictly decreasing in $z_i$. % $\E\big[\varphi_k(x, y)\big] > 0,~\forall k \in [K]$.
    \item $\displaystyle \Psi(z) \,=\, \Psi'(z) / \Psi''(z)$ where $\Psi': \R^K \rightarrow \R_+$ is concave, $\Psi'': \R^K \rightarrow \R_+$ is convex, $\Psi''(z) > 0~ \forall z \in \R^K$, and $\Psi'(z) - M(\theta^*)\Psi''(z)$ is strictly decreasing in each $z_i$.
\end{enumerate}
 Then there exists a weight function $w: \R^D \times \Y \rightarrow \R^m_+$ such that $\theta^*$ is the unique solution for the following weighted loss minimization on the train distribution:
\[
% \underset{\theta \in \R^D}{\argmin}~ M(\theta) \,=\, \underset{\theta \in \R^D}{\argmin}~
\min_{\theta \in \R^D}~\E_{(x,y) \sim \cD'}\Big[\big(w(x, y)\big)^\top\, L\big(h(x; \theta), y\big)\Big].
\]
\end{theorem}

%\comment{TODO: Include conditions on data distribution under which each $\phi_k$ is strictly convex.}
%\hn{TODO: show that this result also holds if the train and test distributions are different.}
%The following result then shows that for test metrics of the suggested form, the MOEW method can find the optimal model 

This theorem shows an equivalence between \textit{maximizing} a metric $M$ of the prescribed form, and \textit{minimizing} an example-weighted loss. We note that similar results have been shown without the strict convexity assumption on $\ell_k$ by making additional assumptions on the data distribution \cite{Koyejo+14} or by allowing the use of stochastic classifiers \citep{Narasimhan:2015b}. %
We also note that  while the above result is stated for general $m$-dimensional losses $L$, for the evaluation metrics considered in this paper, we find it sufficient to work with 1-dimensional losses, and consequently, 1-dimensional weighting functions.

%The monotonicity assumption on $\Psi$ ensures that that the lower the individual losses $\phi_k$, the higher is the metric.
\subsection{Generalization Bound}
To better understand the effect of the embedding dimension and the size of the validation set on the generalization error of the MOEW method, we provide a theoretical analysis of an instance of MOEW where the \emph{GetCandidates} algorithm does an exhaustive search over a fixed candidate set $\mathcal A$ that covers the unit ball.

\begin{theorem}%[\textbf{Generalization bound}]
\label{thm:generic_bound}
Let $\alpha \in \mathds{B}^d$ be the MOEW coefficient vector in a $d$-dimensional unit ball. Let $\cV \sim \mathcal{D}^N$ be a validation set of size $N$.  Denote the empirical versions of the test metric in \eqref{eqn:metric} as: 
\begin{eqnarray*}
\hat M(\theta) &\coloneqq& \Psi(\hat \phi_1(\theta), \ldots, \hat \phi_K(\theta)),\\
\hat \phi_k(\theta) &\coloneqq& \frac 1 N \sum_{(x,y) \in \cV} \ell_k(x,y;\theta).
\end{eqnarray*}
Let $\alpha^* \coloneqq \arg \max_{\alpha \in \mathds{B}^d} M(\hat \theta(\alpha))$ be the optimal coefficient vector in the unit ball, and $\hat \alpha \coloneqq \arg \max_{\alpha \in \mathcal{A}} \hat M(\hat \theta(\alpha))$ be the empirically optimal among a candidate set $\mathcal{A}$. Assume:\\[4pt]
(A) For all $k$, $\ell_k$ is sub-Gaussian with parameter $\sigma$.\\
(B) $\Psi$ is $L_\Psi$-Lipschitz continuous in $\phi's$ w.r.t. the $L_2$ norm.\\
(C) For all $k$, $\phi_k(\hat \theta(\alpha))$ is $L_\phi$-Lipschitz continuous in $\alpha$ w.r.t. the $L_2$ norm.\\[4pt]
For $N \geq 9 \sigma^2 K$, there is a candidate set $\mathcal A$ such that with probability $1-\delta$, $M(\hat \theta(\alpha^*)) - M(\hat \theta(\hat \alpha))$ is bounded by:
\begin{align*}
%M(\hat \theta(\alpha^*)) - M(\hat \theta(\hat \alpha)) \leq \nonumber \\
\frac{\sigma L_\Psi \sqrt K} {\sqrt N} \left( \sqrt{4 d \ln {\frac N {\sigma^2}} + 8 \ln \frac {2K} \delta}  + 3 L_\phi \right) .
\end{align*}
\end{theorem}

Theorem~\ref{thm:generic_bound} establishes that an exhaustive search in the unit ball finds a solution that approaches the optimal test metric at a rate $\tilde O(\sqrt{d \log N /N})$. Note that the Lipschitz constant $L_\phi$ depends on the metric function, model hypothesis space, training loss function, training algorithm and the size of the training set. A more concrete bound can possibly be established with convexity assumption on the loss function and well-behaved hypothesis spaces.

\section{Experimental Results}
\label{sec:exps}

In this section, we illustrate the value of our proposal by comparing it to common strategies on a diverse set of example problems. Unless otherwise noted, for our proposal, we first create a $d$-dimensional embedding of training pairs $\mathcal{T}$ by training an autoencoder that has $d$ nodes in the middle layer. We sample $B\times K$ candidate $\alpha$'s in a $d$-dimensional ball of radius $R$ using GP-BUCB with $p=q=68.3$ and an RBF kernel, whose kernel width was set to be equal to $R$. The noise level of the GP was determined based on the metric noise level of uniform weighting models. For a fair comparison, for competing methods, we also train the same number of models (with random initialization), and pick the one with the best validation metric. Both the autoencoder and the main models were trained for 10k steps using Adam optimizer \citep{Kingma:2015} with learning rate 0.001. We used squared loss for numeric, hinge loss for binary, and cross-entropy loss for multiclass label/features. %The learned weights were batch-normalized during training to stabilize the step size.  
To mitigate the randomness in the result, we repeat the whole process 100 times and report the average and error margin\footnote{The code on public datasets is available at the following GitHub address: https://github.com/google-research/google-research/tree/master/moew.}.

% This comes out of nowhere - do we need it?
%We imposed no regularization on model parameters.

Our experiments are designed to test whether MOEW can help, irregardless of the model structure. Therefore, for simplicity, model architecture and hyperparameters were optimized for each experiment without any weights, then fixed for the weighting experiments. In practice, one might get some additional gains by tuning MOEW and model hyperparameters jointly.

The ability to optimize \emph{any testing metric} is a unique benefit of MOEW over other metric-specific methods. For that reason we focus on demonstrating it does help optimize non-standard metrics and do not consider standard evaluation metrics such as AUC or F-score.

% In this section, we describe the experimental result of our proposal. We compare three training weighting methods: uniform weights, importance weights and our proposal. For our proposal, we first use a three-hidden-layer fully connected autoencoder to project $(X, Y)$ onto a 3-dimensional space, $[0, 1]^3$. Candidate $\alpha$'s are then sampled within a 3-dimensional ball in $N=10$ batches, each with $K=20$ candidate $\alpha$ samples. To make a more fair comparison, for uniform and importance weights, we also train 200 models, and pick the model with the best validation error. To mitigate the randomness in the $\alpha$ sampling and optimization processes, we repeat the whole process 100 times, and report the average test metrics and their 95\% error margins.

\begin{figure}[t]
\centering
\vspace{-3pt}
\begin{subfigure}{.49\linewidth}
    \centering
    \includegraphics[width=\linewidth]{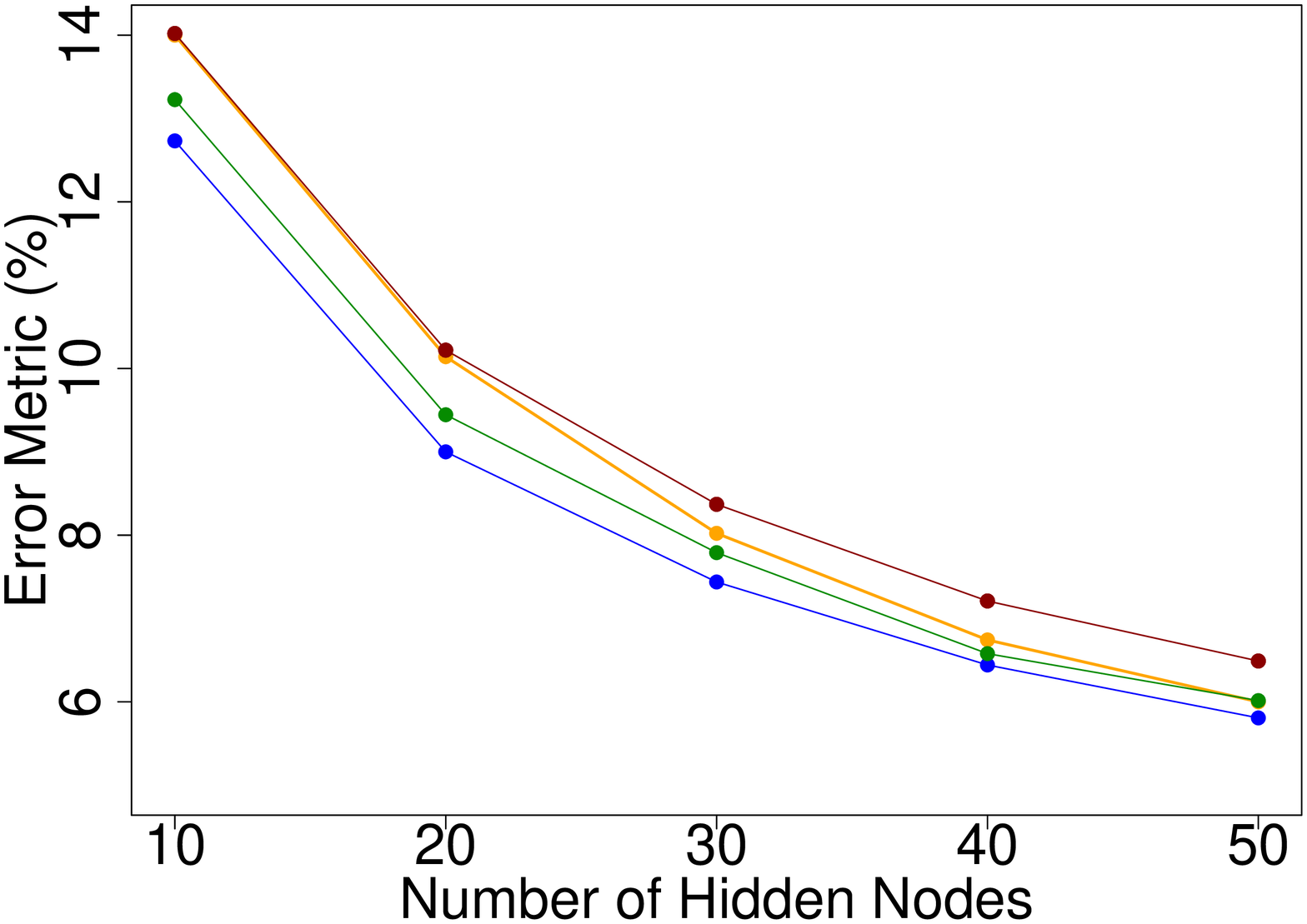}
    \caption{Number of Hidden Nodes}\label{fig:mnist}
\end{subfigure} %
\begin{subfigure}{.49\linewidth}
    \centering
    \includegraphics[width=\linewidth]{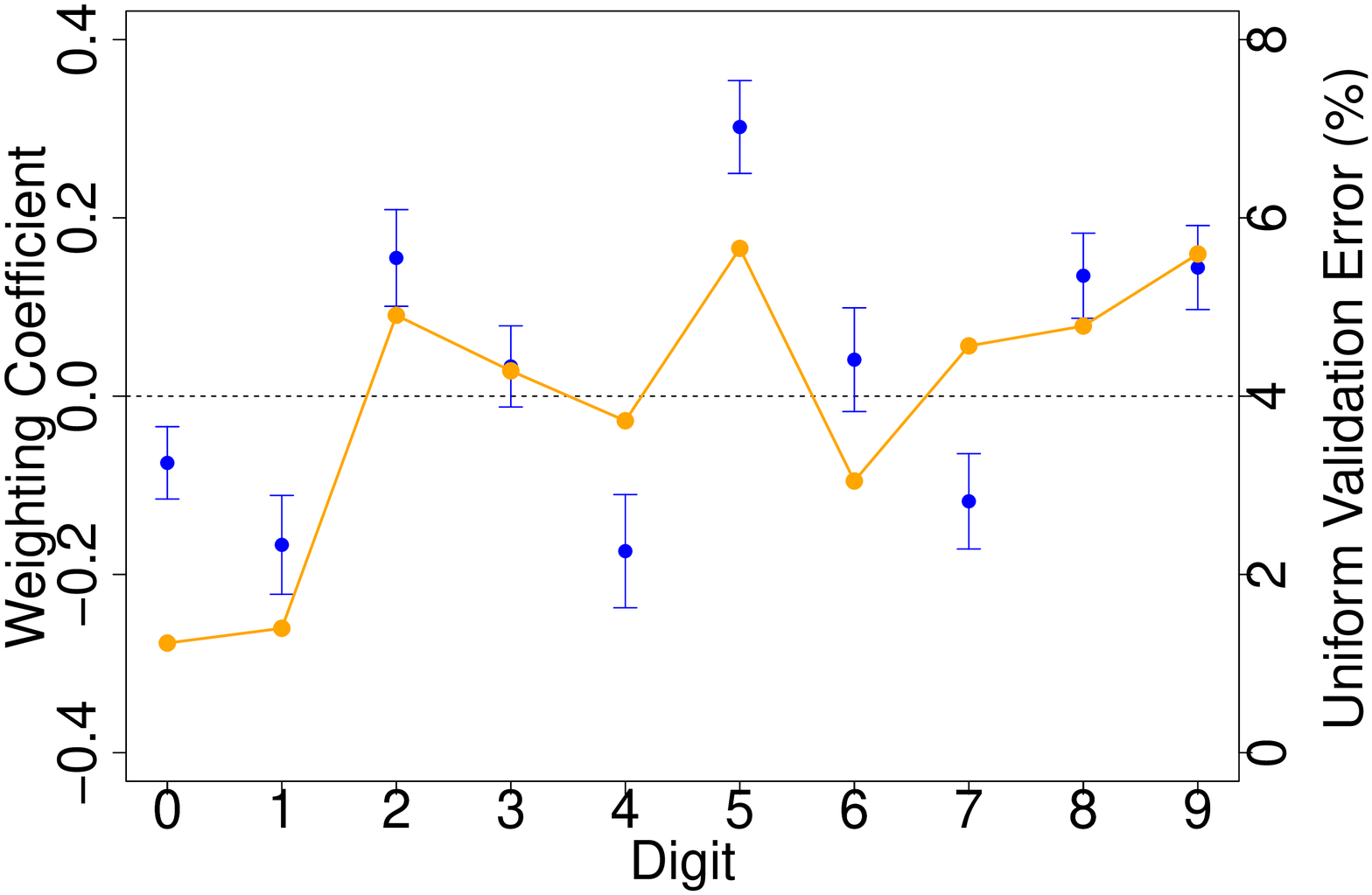}
    \caption{Analysis of Digits}\label{fig:digits}
\end{subfigure} %
\caption{(a) MNIST test metric for varying number of hidden nodes in the main model. We compare classifiers trained with uniform weighting of examples (orange),  random weighting of examples (red), MOEW with random $\alpha$'s (green) and MOEW with GP-BUCB selected $\alpha$'s (blue). The $95\%$ error margin for any point is less than $0.1\%$. (b) With $h=50$ hidden nodes in the main model, average $\alpha$ of each digit and their 95\% error margin for GP-BUCB (blue, axis on the left), as well as the validation error of uniform weighting model (orange, axis on the right).
}
\vspace{-12pt}
\end{figure}

\subsection{MOEW Performance Versus Model Complexity}\label{sec:complexityexps}
We  examine the performance of MOEW in relation to the complexity of the main model and the weighting model.
~\\[-10pt]

\textbf{Main Model Complexity.}\label{sec:MNIST}
MOEW is most valuable when the model class has limited flexibility, and an example weighting strategy is helpful in finding a model that best benefits the test metric.
%If the model is flexible enough, and the training data is clean and sufficient to fit a perfectly accurate model for all parts of the feature space, then MOEW is not needed.  MOEW will be most valuable when the model must take some trade-off, and we can learn an example weighting in a way that best benefits the test metric. 
We illustrate this effect by evaluating MOEW with classifiers of varying complexity. For this experiment, we used the MNIST handwritten digit database \citep{Lecun:2010}. %, and train on it with classifiers of varying complexity.

We used training/validation/test split of sizes 55k/5k/10k respectively. For the MOEW model, to simplify analysis, instead of using an autoencoder to learn the embedding, we directly used the class label as the 10-dimensional embedding. The main classifier was a sigmoid-activation network with $784\rightarrow h\rightarrow 10$ nodes in each layer, with $h\in\{10,20,30,40,50\}$. We used $B=10$ batches of $K=20$ $\alpha$'s, which were either generated randomly or based on GP-BUCB, and compared against the best-of-200 uniform and random weighted models. The error metric was taken to be the maximum of the error rates for each digit: $\max_{i \in \{0, \dots, 9\}} \Pr\{\hat y \neq i |y = i\}$.

Figure~\ref{fig:mnist} shows the test error metric for different model complexities, averaged over 100 runs.  For models of limited complexity, our proposal (blue and green) clearly outperforms a classifier trained with uniform weighting of examples (orange) and a classifier trained with random weighting of examples (red). The benefit is smaller for models that are more flexible. In most real-world situations, the model complexity is limited either by practical constraints on computation, or otherwise to avoid overfitting to smaller training datasets. In such cases there might be an inherent trade-off in the learning process, and we expect MOEW to deliver the most improvement. In addition to the above finding, we observe that candidate $\alpha$'s generated based on GP-BUCB delivers better performance than those generated randomly.

Figure~\ref{fig:digits} shows the average weighting coefficients (i.e., $\alpha$ in \eqref{eq:weightembedding}) of the ten digits generated by GP-BUCB with $h=50$ hidden nodes in the main model (blue). It indicates that MOEW learns to upweight digits 2, 5, 8 and 9, and downweight 0, 1, 4 and 7. Such a result is consistent with the validation error in the uniform weighting model (orange), where digits 2, 5, 8 and 9 have the highest error.
~\\[-10pt]

\textbf{MOEW Model Complexity and Risk of Overfitting.}\label{sec:wine}
In this experiment, we study the effect of the embedding dimension $d$, the number of batches $B$ and the exploration-exploitation parameters $p$ and $q$ on the MOEW performance. Specifically, we are interested in the difficulty for GP-BUCB to find the optimal solution, as well as the risk of overfitting to the validation dataset. We used the wine reviews dataset from Kaggle (www.kaggle.com/zynicide/wine-reviews). The task is to predict the price of the wine using 39 Boolean features describing characteristic of the wine and the quality score (points), for a total of 40 features. We calculate the error in percentage of the correct price, and want the model to have good accuracy across all price ranges. To that end, we set the test metric to be the worst of the errors among 4 quartiles $\{q_i\}$ of the price (thresholds are 17, 25 and 42 dollars):
$\max_{i \in \{0, \dots, 3\}} E_{\text{price}(x) \in q_i}[|\hat y/y - 1|]$.

We used training/validation/test split of sizes 85k/12k/24k respectively. We applied a log transformation to the label (price) and used mean squared error as the training loss on the log-transformed prices (for all weightings).  For MOEW, we used a $d$-dimensional embedding, created by training a sigmoid-activation autoencoder network on the $(x,y)$ pair, with $41\rightarrow 100\rightarrow d\rightarrow 100\rightarrow 41$ nodes in each layer. The main regressor was a sigmoid-activation network with $40\rightarrow 20\rightarrow 10\rightarrow 1$ nodes in each layer. We used $B$ batches of $K=20$ $\alpha$'s in our proposed method. 

\begin{figure}[t]
\centering
\begin{subfigure}{.49\linewidth}
    \centering
    \includegraphics[width=\linewidth]{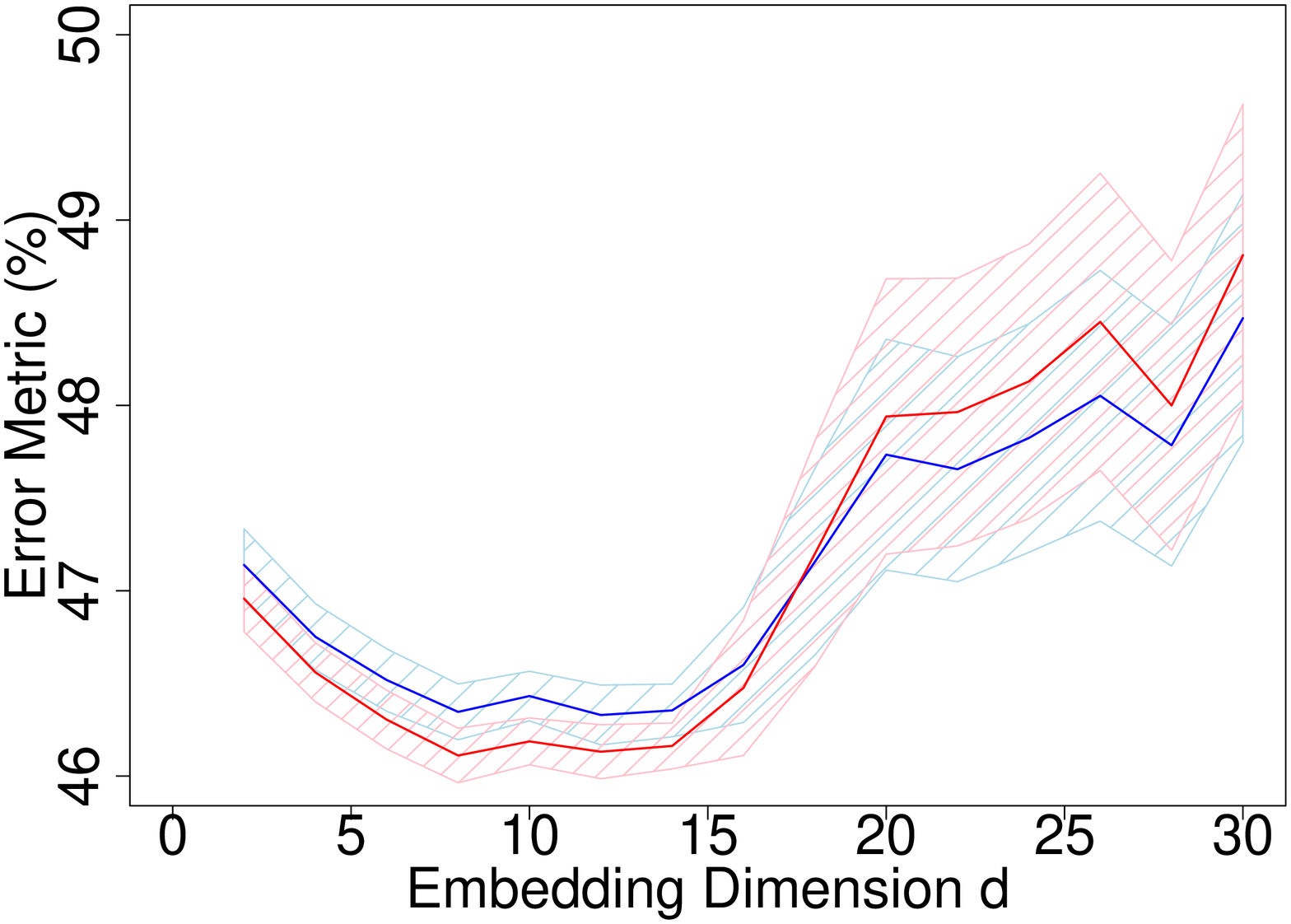}
    \caption{Embedding Dimension $d$}\label{fig:embeddingdim}
\end{subfigure} %
\begin{subfigure}{.49\linewidth}
    \centering
    \includegraphics[width=\linewidth]{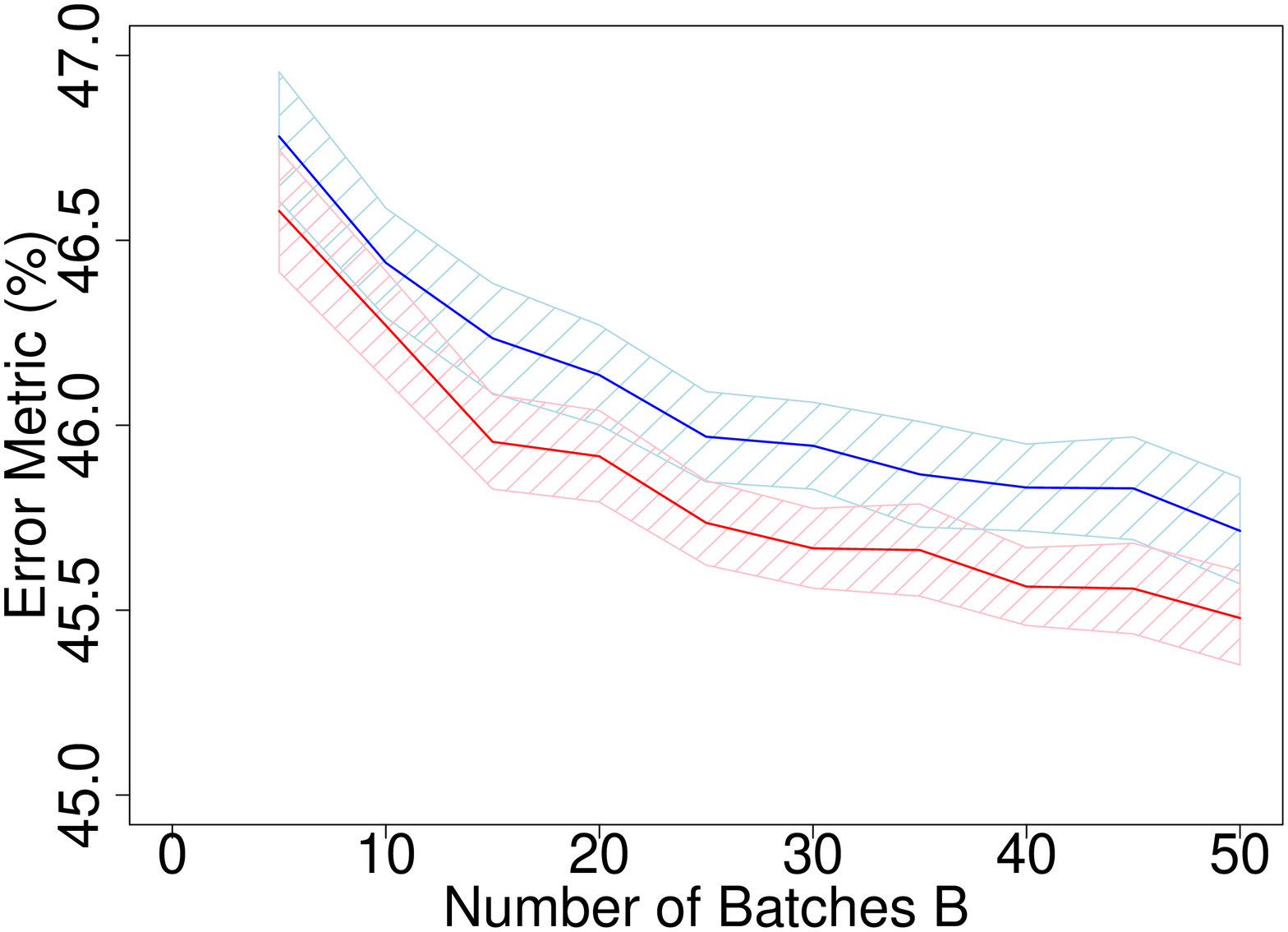}
    \caption{Number of Batches $B$}\label{fig:numbatches}
\end{subfigure} %
\caption{Metrics for validation (red) and test (blue) on wine data with (a) embedding dimension $d\in\{2,4,\dots, 30\}$ and number of batches $B=10$, and (b) $d=10$  and $B\in\{5,10,\dots, 50\}$. The pink and light blue shaded areas indicate the $95\%$ confidence bands.
}
\vspace{-10pt}
\end{figure}

In the first study, shown in figure \ref{fig:embeddingdim}, we fixed the number of batches $B=10$ and varied the embedding dimension $d\in\{2,4,\dots, 30\}$. The result indicates that the test performance of MOEW improves as we increase the embedding dimension up to around $d=8$, at which point the test error metric is $46.33 \pm 0.16$. As a comparison, the best-of-200 uniform weighted models achieves an average test error metric of $52.00 \pm 0.31$. On the other hand, the result also suggests that the validation performance begins to drop when $d>14$, which indicates that GP-BUCB was unable to converge to good solutions in such high-dimensional spaces with 200 candidate $\alpha$'s. In addition, the test-metric-to-validation-metric-ratio is slightly larger when $d$ is small, which suggests that there might be overfitting to the validation set with \emph{small} embedding dimension $d$. This is in fact intuitive: in a high-dimensional space with a limited number of $\alpha$'s, GP-BUCB acts similarly to pure exploration. Because there is less exploitation, there is also less overfitting.

In the second study, shown in figure \ref{fig:numbatches}, we fixed the embedding dimension $d=10$ and investigated the performance of MOEW with number of batches $B\in\{5,10,\dots,50\}$. The figure indicates that both the validation and test performance improves as we sample more batches of candidate weighting parameters. In addition, the gap between validation and test metrics does not widen, suggesting that the risk of overfitting is small in the range of values we experimented with. As a comparison with the $B=50$ case, the best-of-1000 uniform model achieves test performance $48.27 \pm 0.41$.

In the third study, we fixed the embedding dimension $d=10$ and number of batches $B=10$, and study the effect of $p$ and $q$ on MOEW performance. When $p=q=0$, $75$, $95$, $99.99$ and $99.999$, MOEW test errors are $47.07\pm0.19$, $46.32\pm0.15$, $46.28\pm0.15$, $46.28\pm0.16$ and $46.46\pm0.17$, respectively. MOEW is insensitive to $p$ or $q$ if exploration is reasonably large.

\subsection{MOEW Performance on Small Data}\label{sec:crime}
In this example, we examine MOEW on a small dataset, namely the Communities and Crime dataset from the UCI Machine Learning Repository~\citep{UCI:2017}, which contains the violent crime rate of 1994 communities. The goal is to predict whether a community has violent crime rate per 100k population above 0.28.

In addition to obtaining an accurate classifier, we also aim to improve its \emph{fairness}. To this end, we divide the communities into 4 groups based on the quartiles of white population percentage in each community (thresholds are 63\%, 84\% and 94\%). We seek a classifier with high accuracy, but that has similar false positive rates (FPR) across racial groups. We evaluate classifiers based on two metrics: overall accuracy and the difference between the highest and lowest FPR across four racial groups (fairness violation).

We used a linear classifier with 95 features, including the percentage of African American, Asian, Hispanic and white population. For MOEW, those features and the label were projected to a 4-dimensional space using an autoencoder with $96\rightarrow 10\to4\rightarrow10\rightarrow96$ nodes. We sampled candidate $\alpha$'s in $B=10$ batches of size $K=5$.

In practice, there is usually a trade-off between accuracy and fairness \citep[see, e.g.,][]{Hardt:2016, Goh:2016}. To explore this trade-off, we consider two approaches. The first is a vanilla approach, where we use MOEW to minimize the difference between the highest and lowest FPR across the 4 groups (i.e., minimize fairness violation), with an identical decision threshold for all groups (chosen to maximize overall accuracy). The second is the post-shift approach \citep{Hardt:2016}, after training the model, we use MOEW to maximize accuracy with separate decision thresholds for each group chosen to achieve the same FPR on the training data while maintaining the same overall coverage. 

In the first study, we sample 994/500/500 training/validation/testing examples purely randomly, and compare MOEW with the best-of-50 uniform weighted model. The results are summarized in Table \ref{tab:Crime}. With the vanilla approach, MOEW reduces fairness violation by over 20\% and yet achieves the same accuracy compared to uniform weighting. With the post-shift approach, MOEW improves accuracy and reduces fairness violation at the same time.

\begin{table}[t]
\centering
\caption{Average accuracy and fairness violation with their 95\% error margin with IID training/test split. We consider (Vanilla) minimizing fairness violation with identical decision threshold across racial groups, and (Post-Shift) maximizing accuracy with racial-group-specific thresholds for equal FPR on the training data.}
\vspace{-0.5em}
\label{tab:Crime}
\begin{tabular}{lcccc}
\toprule
& \multicolumn{2}{c}{Vanilla Approach} \\
\cmidrule(r){2-3}
 & Uniform & MOEW     \\
\midrule
Accuracy (\%) & $86.82 \pm 0.07$    & $86.84\pm0.12$\\
Fairness Violation (\%) & $79.23 \pm 2.35$    & $57.74\pm4.64$\\
\midrule
& \multicolumn{2}{c}{Post-Shift Approach} \\
\cmidrule(r){2-3}
 & Uniform & MOEW     \\
\midrule
Accuracy (\%) & $79.12 \pm 0.12$    & $79.95\pm0.17$\\
Fairness Violation (\%) & $12.26 \pm 0.30$    & $10.06\pm0.54$\\
\bottomrule
\end{tabular}
\vspace{-1em}
\end{table}

In the second study, we sample 994/500/500 training/validation/testing examples such that in the training data, 76\% of the communities have a population count above the median population, whereas in the validation/testing set, 40\% of the communities have a population count above the median. We compare MOEW with the best-of-50 optimal domain adaptation training weights provided by \citet{Shimodaira:2000}. The results are summarized in Table \ref{tab:Crime2}. MOEW performed similarly to \citet{Shimodaira:2000} with the vanilla approach, and better in fairness violation with post-shift.

\begin{table}[t]
\centering
\caption{Average accuracy and fairness violation with their 95\% error margin with non-IID training/test split.}
\vspace{-0.5em}
\label{tab:Crime2}
\begin{tabular}{lcccc}
\toprule
& \multicolumn{2}{c}{Vanilla Approach} \\
\cmidrule(r){2-3}
 & Shimodaira & MOEW     \\
\midrule
Accuracy (\%) & $85.59 \pm 0.05$    & $85.59\pm0.06$\\
Fairness Violation (\%) & $37.53 \pm 0.40$    & $37.78\pm0.39$\\
\midrule
& \multicolumn{2}{c}{Post-Shift Approach} \\
\cmidrule(r){2-3}
 & Shimodaira & MOEW     \\
\midrule
Accuracy (\%) & $72.46 \pm 0.11$    & $72.33\pm0.12$\\
Fairness Violation (\%) & $6.63 \pm 0.44$    & $4.61\pm0.41$\\
\bottomrule
\end{tabular}
\vspace{-1em}
\end{table}

\subsection{Spam Blocking}\label{sec:spam}
For this problem from a large internet services company, the goal is to classify whether a result is spam, and this decision affects whether the result provider receives ads revenue from the company. Thus, it is more important to block more expensive spam results, but it is also important not to block any results that are not spam, especially results with many impressions. We used a simplified test metric that captures these different goals (the actual metric is more complex and proprietary).  Specifically, for each method we set the classifier decision threshold so that 5\% of the validation set is labelled as spam. We then sum the costs saved by blocking correctly identified spams and divide it by the total number of blocked impressions of incorrectly-identified spams.

The datasets contain 12 features. We trained an autoencoder on the 12 features plus label, with layers of $13 \rightarrow 100 \rightarrow 3 \rightarrow 100 \rightarrow 13$ nodes, and used the middle layer as an embedding. For each weighting method, we built a sigmoid-activation network classifier with architecture $12 \rightarrow 20 \rightarrow 10 \rightarrow 1$. Candidate $\alpha$'s were sampled $K=20$ at a time in $B=10$ rounds of sampling.    

In the first study, we divide the dataset such that the 180k training dataset is 25\% spam, and is not IID with the validation/test datasets, which are IID and have 10k/30k examples with 5\% spam. In the second study, we divide the dataset such that half of examples are from large result providers in the 180k training dataset, and 20\% of examples are from large result providers in the 10k/30k validation/test datasets.

Table~\ref{tab:Spam} compares MOEW with two common choices in practice: uniform weighting and optimal domain adaptation weighting \citep{Shimodaira:2000}. We have normalized the reported scores so that the average uniformly weighted test metric is 1.0. Our proposed method clearly outperforms both uniform and optimal domain adaptation weighting.

\begin{table}[t]
\centering
\caption{Average test metric and 95\% error margin for Spam Blocking with models trained with uniform weighting, optimal domain adaptation weighting (Shimodaira) and MOEW. Larger test metric is better.}
\vspace{-0.6em}
\label{tab:Spam}
\begin{tabular}{lc|c}
\toprule
 & Study 1 & Study 2     \\
\midrule
Uniform & $1.000\pm 0.040$   &   $1.000 \pm 0.124$  \\
Shimodaira & $1.210\pm 0.063$ & $1.010 \pm 0.137$  \\
MOEW & $1.849\pm0.133$ & $8.057\pm0.923$ \\
\bottomrule
\end{tabular}
\vspace{-1em}
\end{table}

\subsection{Web Page Quality}\label{sec:DSA}
This binary classifier example is from a large internet services company. The goal is to identify high quality webpages, thresholded such that 40\% of examples are classified as high quality. The company performed several rounds of human ratings of example web pages. In the early rounds, the label was binary (high/low quality). Later, human raters provided finer-grained labels, scoring the quality in $[0, 1]$. The test metric for this problem is the mean fine-grained quality score of the positively classified test examples.

We trained a six-feature sigmoid-activation classifier with $6\rightarrow20\rightarrow 10\rightarrow 1$ nodes on the 62k examples with binary labels. To fit MOEW, we trained an autoencoder that mapped the six features plus label onto a 3-dimensional space: $7\rightarrow100\rightarrow3\rightarrow100\rightarrow7$. We sampled $K=20$ candidate $\alpha$'s for $B=10$ rounds. 

The validation data and test data each have 10k web pages labeled with a fine-grained score in $[0,1]$. The datasets are not IID: the training data is the oldest data, the test data is the newest data, with validation data in-between. The average quality score of 40\% selected web pages is $0.7113 \pm 0.0004$ with uniform weighting, and $0.7176\pm0.0004$ with MOEW. Note that in this example, domain adaptation weighting results in uniform weighting.

\section{Conclusions}
\label{sec:con}
We proposed learning example weights to sculpt a standard loss function into one that better optimizes a test metric for the test distribution. We demonstrated substantial benefits on public benchmark datasets and real-world applications, for problems with non-IID train/test distributions and custom metrics that incorporated multiple objectives. 

To limit the amount of validation data needed, we define the example weighting model over a low-dimensional space. Our experiments used the low-dimensional embedding of $(x, y)$  produced by an autoencoder, but we hypothesize that a discriminatively-trained embedding could be more optimal, or that using a small subset of semantically-meaningful features could be more interpretable.

We hypothesize that the MEOW could also be useful for other purposes. For example, they may be useful for guiding active sampling, suggesting one should sample more examples in feature regions with high training weights. And we hypothesize one could downsample areas of low-weight to reduce the size of training data for faster training and iterations, without sacrificing test performance.

\bibliographystyle{icml2019}
\bibliography{references}
\newpage

\section{Appendix}

\subsection{Proof of Theorem 1}

We consider a learning problem where the goal is to maximize a metric $M(\theta)$  defined on a distribution $D$ over the space of model parameters:
\begin{equation}
\theta^* \in \argmax_{\theta \in \R^D}\, M(\theta).
\label{eq:learning-problem}
\end{equation}
We will denote:
\begin{eqnarray}
\R_+ &=& \{x \in \R \,|\, x \geq 0\}, \nonumber \\
\R_{++} &=& \{x \in \R \,|\, x > 0\}. \nonumber
\end{eqnarray}

\begin{proof}
The proof is based on techniques from \citet{Narasimhan:2015b} used to develop algorithms for optimizing similar families of evaluation metrics. While the focus in \citet{Narasimhan:2015b} was on developing optimization algorithms for non-decomposable evaluation metrics, we extend their techniques to provide a characterization of the optimal model for these metrics.

\paragraph{Proof for Family 1.} We will find it useful to first re-write $\Psi$ in terms of its concave conjugate. The concave conjugate of $\Psi$ for dual variables $\beta \in \R^K$ is given by:
\begin{eqnarray*}
\Psi^*(\beta) \,=\, \min_{z \in \R^K} \big\{ -\beta^\top z \,-\, \Psi(z) \big\}.
\end{eqnarray*}
Since we assume that $\Psi$ is concave in $z$, by standard results its conjugate $\Psi^*$ is concave in $\beta$. Since we also assume that $\Psi$ is strictly decreasing in $z$ and we have negated the dual variables in the conjugate definition, $\Psi^*$ is finite only for positive dual variables $\beta \in \R_{++}^K$. The concavity of $\Psi$ then allows us to re-write $\Psi$  in terms of its conjugate, giving us:
\begin{eqnarray}
\Psi(z) \,=\, \min_{\beta \,\in\, \R_{++}^K} \big\{ -\beta^\top z \,-\, \Psi^*(\beta) \big\}
\label{eq:psi-rewritten}
\end{eqnarray}

The proof then follows by using \eqref{eq:psi-rewritten} to re-write the evaluation metric $M$ as a minimum over the dual space, and re-write the learning problem in \eqref{eq:learning-problem} as a min-max optimization involving a minimization over dual variables %(or equivalently over weight functions) 
and a maximization over $\theta$. 

%\allowdisplaybreaks

Rewriting $M$, we get:
\begin{eqnarray}
M(\theta)
&=& \Psi\big(\phi_1(\theta), \ldots, \phi_K(\theta)\big)\\
&=&\min_{\beta \in \R_{++}^K}\bigg\{
\underbrace{
  \sum_{k=1}^K -\beta_k \phi_k(\theta) \,-\, \Psi^*(\beta)
}_{
  F(\theta, \beta)
}
\bigg\}.
\label{eq:M-rewritten}
\end{eqnarray}
Note that the inner objective in \eqref{eq:M-rewritten} is concave in $\theta$ for any fixed $\beta$ (as each $-\phi_k$ is concave), and convex in $\beta$ for a fixed $\theta$. We further have:
\begin{eqnarray}
\lefteqn{
F(\theta, \beta)
}\nonumber\\
&=& \sum_{k=1}^K -\beta_k 
\E_{(x,y)\sim \cD}\bigg[\big(\varphi_{k}(x, y) \big)^\top\, L\big(h(x; \theta), y\big)\bigg]\nonumber\\
&&\hspace{6cm} - \Psi^*(\beta) \nonumber\\
&=& -\E_{(x,y)\sim \cD}\bigg[\bigg(\sum_{k=1}^K \beta_k \varphi_{k}(x, y)\bigg)^\top L\big(h(x; \theta), y\big)\bigg]\nonumber\\
&&\hspace{6cm} - \Psi^*(\beta) \nonumber\\
&=& -\E_{(x,y)\sim \cD}\Big[\big(\bar{w}(x, y; \beta)\big)^\top  L\big(h(x; \theta), y\big)\Big] - \Psi^*(\beta).
\nonumber\\
\label{eq:F}
\end{eqnarray}
where $\bar{w}: \R^D \times \Y \rightarrow \R_+^n$ is defined as $\bar{w}(x, y; \beta) \,=\, \sum_{k=1}^K \beta_{k}\, \varphi_{k}(x, y)$.

Thus the learning problem in \eqref{eq:learning-problem} can be equivalently written as a max-min optimization problem:
\begin{eqnarray}
\max_{\theta \in \R^D}\,M(\theta) &=&
\max_{\theta \in \R^D}\min_{\beta \in \R_{++}^K}\,F(\theta, \beta).
\label{eq:minmax}
\end{eqnarray}
Applying the min-max theorem (recall that $F$ is concave in $\theta$ and convex in $\beta$):
%we can further re-write \eqref{eq:learning-problem} as a a max-min optimization problem:
% \begin{eqnarray}
% \min_{\theta \in \R^D}\,M(\theta)  &=& \max_{\beta \in \R_{++}^K}\bigg\{\min_{\theta \in \R^D}\,\E\Big[w_\beta(x, y)\, L\big(h(x; \theta), y\big)\Big] \,-\, \Psi^*(\beta)\bigg\}.
% \label{eq:maxmin}
% \end{eqnarray}
\begin{eqnarray}
\max_{\theta \in \R^D}\min_{\beta \in \R_{++}^K}\, F(\theta, \beta) &=& \min_{\beta \in \R_{++}^K}\max_{\theta \in \R^D}\,  F(\theta, \beta).
\label{eq:maxmin}
\end{eqnarray}

Let $\theta^*$ be an optimal solution to the learning problem in \eqref{eq:learning-problem}. 
Then 
$\theta^* \in \argmax_{\theta \in \R^D} \min_{\beta \in \R^K_{++}}\, F(\theta, \beta)$ is a solution to the max-min problem.  
Let $\beta^* \in \argmin_{\beta \in \R^K_{++}}\max_{\theta \in \R^D}\, F(\theta, \beta)$ be an optimal solution to the min-max problem. It is easy to derive from \eqref{eq:maxmin}:
\begin{eqnarray*}
\max_{\theta \in \R^D}\, F(\theta, \beta^*) ~=~ F(\theta^*, \beta^*) ~=~ \min_{\beta \in \R_{++}^K}\,  F(\theta^*, \beta).
\end{eqnarray*}
%%%%
\if 0
\begin{eqnarray*}
\min_{\theta \in \R^D}\, F(\theta, \beta^*) ~=~ \max_{\beta \in \R_{++}^K}\,  F(\theta^*, \beta) ~\geq~ F(\theta^*, \beta^*) ~\geq~ \min_{\theta \in \R^D}\, F(\theta, \beta^*).
\end{eqnarray*}
\fi
%%%%
It follows from the above equality that $\theta^*$ is a maximizer for $F(\theta, \beta^*)$, and hence from \eqref{eq:F}, a maximizer for following (negative) cost-weighted loss with weights $\bar{w}(x, y; \beta^*)$:
\begin{equation}
\max_{\theta \in \R^D}\, -\E_{(x,y)\sim \cD}\big[\big(\bar{w}(x, y; \beta^*)\big)^\top L\big(h(x; \theta), y\big)\big].
\end{equation}
or a minimizer for:
\begin{equation}
\min_{\theta \in \R^D}\, \E_{(x,y)\sim \cD}\big[\big(\bar{w}(x, y; \beta^*)\big)^\top L\big(h(x; \theta), y\big)\big].
\label{eq:weighted-loss}
\end{equation}
Further, since each $\beta^*_k > 0$ and each $\phi_k$ is strictly convex in $\theta$, the weighted loss $\E\big[\big(w(x, y;\beta^*) \big)^\top L\big(h(x; \theta), y\big)\big] \,=\, \sum_{k=1}^K \beta_k^* \phi_k(\theta)$ is also strictly convex in $\theta$ and has a unique minimizer. Hence $\theta^*$ is a unique solution for the weighted loss minimization in \eqref{eq:weighted-loss}.

The above cost-weighted loss is defined on the test distribution $\cD$. It remains to be shown that $\theta^*$ is a unique minimizer for a weighted loss minimization problem on the training distribution $\cD'$.
Let $f, f': \R^D \times \Y \rightarrow \R_+$ denote the probability density functions associated with the test distribution $\cD$ and train distributions $\cD'$ respectively. By our assumption, both distributions have the same support. Letting $A \subseteq  \R^D \times \Y$ denote their support, we have $f(x, y) > 0$ and $f'(x, y) > 0$ for all $(x, y) \in A$. Clearly, for any $\theta$:
\begin{eqnarray*}
\lefteqn{\E_{(x,y)\sim \cD}\big[\big(\bar{w}(x, y; \beta^*)\big)^\top L\big(h(x; \theta), y\big)\big]}\\
&=& \E_{(x,y)\sim \cD'}\bigg[\frac{f(x, y)}{f'(x, y)}\big(\bar{w}(x, y; \beta^*)\big)^\top L\big(h(x; \theta), y\big)\bigg].
\end{eqnarray*}
Thus for $w(x, y; \beta^*) \,=\, \frac{f(x, y)}{f'(x, y)}\,\bar{w}(x, y; \beta^*)$, $\theta^*$ is also a unique solution for the following weighted loss minimization on the training distribution:
\begin{equation*}
\min_{\theta \in \R^D}\, \E_{(x,y)\sim \cD'}\big[\big(w(x, y; \beta^*)\big)^\top L\big(h(x; \theta), y\big)\big].
\end{equation*}

\paragraph{Proof for Family 2.} 
We next move to the second family of evaluation metrics $M$ with a fractional-concave $\Psi$. We will first show that for any metric $M$ in this family there exists an evaluation metric in Family 1 whose optimal solutions are the same as that of $M$. We will then invoke the result for Family 1 to complete the proof. 

%Let $t^* = \min_{\theta \in \R^D}\, M(\theta)$. % and $\theta^*$ denote an optimal solution at which this value is achieved. 
The fractional structure of the evaluation metric gives us that for any $t \in \R_+$:
%\begin{equation}
%M(\theta) \geq t, ~\forall \theta \in \R^D
%~~~\iff~~~
%    \Psi'\big(\phi_1(\theta), \ldots, \phi_K(\theta)\big)
%        ~-~
%    t\,\Psi''\big(\phi_1(\theta), \ldots, \phi_K(\theta)\big) 
%        ~\geq~ 0, ~\forall \theta \in \R^D.
%\label{eq:implication1}
%\end{equation}
\begin{align}
&\forall \theta \in \R^D: M(\theta) \leq t
~\iff~ \nonumber \\
&\forall \theta \in \R^D:
\Psi'(\phi_{1:K}(\theta)) - t \Psi''(\phi_{1:K}(\theta)) \leq 0,
\label{eq:implication1}
\end{align}
and for a fixed $\theta \in \R^D$:
%\begin{align}
%M(\theta) = t
%~~~\iff~~~ \nonumber \\
%    \Psi'\big(\phi_1(\theta), \ldots, \phi_K(\theta)\big)
%        ~-~
%    t\,\Psi''\big(\phi_1(\theta), \ldots, \phi_K(\theta)\big) 
%        ~=~ 0.
%\label{eq:implication2}
%\end{align}
\begin{align}
M(\theta) = t
~\iff~ 
\Psi'(\phi_{1:K}(\theta)) - t \Psi''(\phi_{1:K}(\theta)) = 0.
\label{eq:implication2}
\end{align}

Combining \eqref{eq:implication1} and \eqref{eq:implication2} with $t^* = M(\theta^*)$, we
have that $\theta^* \in \argmax_{\theta \in \R^K} M(\theta)$ \textit{iff}:
% \begin{eqnarray}
% \Psi'\big(\phi_1(\theta), \ldots, \phi_K(\theta)\big)
%         ~-~
%     t^*\,\Psi''\big(\phi_1(\theta), \ldots, \phi_K(\theta)\big) 
%         ~\geq~ 0, ~~~\forall \theta \in \R^K
% \label{eq:geq0}\\
% ~~~~~~\text{  and  }~~
% \Psi'\big(\phi_1(\theta^*), \ldots, \phi_K(\theta^*)\big)
%         ~-~
%     t^*\,\Psi''\big(\phi_1(\theta^*), \ldots, \phi_K(\theta^*)\big) 
%         ~=~ 0,
% \label{eq:eq0}
% \end{eqnarray}
% Combining \eqref{eq:geq0} and \eqref{eq:eq0}, we have that $\theta^* \in \argmin_{\theta \in \R^K} M(\theta)$ \textit{if and only if}:
\begin{equation}
\theta^* ~\in~ \underset{\theta \in \R^K}{\argmax}~\Big\{
\underbrace{
\Psi'(\phi_{1:K}(\theta)) - t^* \Psi''(\phi_{1:K}(\theta))
}_{G(\theta)}
\Big\}.
\label{eq:Gtheta}
\end{equation}

Notice that by our assumption that $\Psi'$ is concave and $\Psi''$ is convex, and since $t^* = \Psi'(\theta^*) / \Psi''(\theta^*) > 0$, $\Psi'(z) - t^*\Psi''(z)$ is concave in $z$. We also assume that $\Psi'(z) - t^*\Psi''(z)$ is strictly decreasing in each $z_j$. Thus $G$ satisfies the assumptions for Family 1. Invoking the theorem statement for Family 1, we have from \eqref{eq:Gtheta} that for any $\theta^* \in \argmax_{\theta \in \R^K} M(\theta)$, there exists 
  a weight function $w: \R^D \times \Y \rightarrow \R^n$ such that $\theta^*$ is the unique minimizer for the weighted loss:
\[
% \underset{\theta \in \R^D}{\argmin}~ M(\theta) \,=\, \underset{\theta \in \R^D}{\argmin}~
\min_{\theta \in \R^D}~\E_{(x,y)\sim \cD'}\big[w^\top(x, y) L\big(h(x; \theta), y\big)\big].
\]
\end{proof}

\subsection{Proof of Theorem 2}

\begin{proof}
Define:
\begin{eqnarray}
\epsilon & \coloneqq & \frac{3 \sigma}{\sqrt{N}},\\
\Delta   & \coloneqq & \frac{\sigma}{\sqrt N} \sqrt{2d \ln(3/\epsilon) + 2\ln (2K/\delta)}.
\end{eqnarray}

Using Hoeffding's inequality \cite{chernoff1952measure}  we have for any fixed $\alpha$ and $k$:
\begin{eqnarray}
\!\!\!\!\! P\{ | \hat \phi_k(\hat \theta(\alpha)) - \phi_k(\hat \theta(\alpha)) | > \Delta \} 
&\leq& 2 e^{-\frac{N\Delta^2}{2\sigma^2}} \\
&=& \frac{\delta}{K (3/\epsilon)^d}
\end{eqnarray}
Let $\mathcal{A}$ be an $\epsilon$-cover of $\mathds{B}^d$ (i.e. $\forall \alpha \in \mathds{B}^d \, \exists \alpha' \in \mathcal{A} : \|\alpha - \alpha'\| \leq \epsilon$). By Lemma 5.2 of \cite{eldar2012compressed} we have $|\mathcal{A}| \leq (1+2/\epsilon)^d$. By the assumption of the theorem we have $\epsilon \leq 1$ and thus $|\mathcal{A}| \leq (3/\epsilon)^d$. Using the above and union bounding over all elements of $\mathcal{A}$ and all $k$, we have:
\begin{equation}
P\{\forall k \, \forall \alpha \in \mathcal{A}: | \hat \phi_k(\hat \theta(\alpha)) - \phi_k(\hat \theta(\alpha)) | \leq \Delta \}  \geq 1 - \delta.
\end{equation}
Since $\Psi$ is Lipschitz continuous we get:
\begin{equation}
P\{\alpha \in \mathcal{A}: | \hat M(\hat \theta(\alpha)) - M(\hat \theta(\alpha)) | \leq L_\Psi \sqrt K \Delta \}  \geq 1 - \delta. \label{eqn:cover_bound}
\end{equation}
Therefore with probability no less than $1 - \delta$ we have:
\begin{eqnarray}
\!\!\!\!\!\!\!\! \forall \alpha' \in \mathcal{A}: M(\hat \theta(\hat \alpha))
&\geq& \hat M(\hat \theta(\hat \alpha)) - L_\Psi \sqrt K \Delta \label{eqn:alpha_hat_in_a} \\
&\geq& \hat M(\hat \theta(\alpha')) - L_\Psi \sqrt K \Delta \label{eqn:alpha_hat_optimum} \\
&\geq& M(\hat \theta(\alpha')) - 2 L_\Psi \sqrt K \Delta \label{eqn:alpha_prime_in_a}
\end{eqnarray}
Lines \eqref{eqn:alpha_hat_in_a} and \eqref{eqn:alpha_prime_in_a} use the bound in \eqref{eqn:cover_bound} (since $\hat \alpha \in \mathcal{A}$), and line \eqref{eqn:alpha_hat_optimum} is by the definition of $\hat \alpha$. Since $\mathcal{A}$ is an $\epsilon$-cover of $\mathds{B}^d$, using $(L_\Psi L_\phi \sqrt K)$-Lipschitz continuity of $\Psi$ w.r.t. $\alpha$ we get:
\begin{equation}
\forall \alpha \in \mathds{B}^d \!\!: M(\hat \theta(\hat \alpha))
\!\geq\! M(\hat \theta(\alpha)) - 2 L_\Psi \sqrt K \Delta - L_\Psi L_\phi \sqrt K \epsilon,
\end{equation}
which gives out the bound after substitution of $\Delta$ and $\epsilon$.
\end{proof}

\end{document}